\pdfoutput=1

\documentclass[11pt]{article}

\usepackage[]{acl}

\usepackage{times}
\usepackage{latexsym}

\usepackage[T1]{fontenc}

\usepackage[utf8]{inputenc}

\usepackage{microtype}

\usepackage{graphicx}
\usepackage{caption}
\usepackage{subcaption}
\usepackage{diagbox}

\title{Punctuation Restoration in Spanish Customer Support Transcripts using Transfer Learning}

\author{Xiliang Zhu, \ Shayna Gardiner, \ David Rossouw  \\ {\bf Tere Roldán, \ Simon Corston-Oliver}\\
        Dialpad Canada Inc.\\
        \texttt{\{xzhu, sgardiner, davidr\}@dialpad.com} \\
        \texttt{\{tere.roldan, scorston-oliver\}@dialpad.com} \\}

\begin{document}
\maketitle
\begin{abstract}
Automatic Speech Recognition (ASR) systems typically produce unpunctuated transcripts that have poor readability. In addition, building a punctuation restoration system is challenging for low-resource languages, especially for domain-specific applications. In this paper, we propose a Spanish punctuation restoration system designed for a real-time customer support transcription service. To address the data sparsity of Spanish transcripts in the customer support domain, we introduce two transfer-learning-based strategies: 1) domain adaptation using out-of-domain Spanish text data; 2) cross-lingual transfer learning leveraging in-domain English transcript data. Our experiment results show that these strategies improve the accuracy of the Spanish punctuation restoration system.
\end{abstract}

\section{Introduction}
\label{section:1}

Automatic Speech Recognition (ASR) systems play an increasingly important role in our daily lives, with a wide range of applications in different domains such as voice assistant, customer support and healthcare. However, ASR systems usually generate an unpunctuated word stream as the output. Unpunctuated speech transcripts are difficult to read and reduce overall comprehension \cite{inproceedings-jones} . Punctuation restoration is thus an important post-processing task on the output of ASR systems to improve general transcript readability and facilitate human comprehension.

Punctuation restoration for transcripts of Spanish-speaking customer support telephone dialogue is a non-trivial task. First, real-world human conversation transcripts have unique characteristics compared to common written text, e.g., filler words and false starts are common in spoken dialogue. Moreover, further challenges arise when addressing noisy ASR transcripts in a specific domain, as the lexical data distribution can be quite different compared to public Spanish datasets. Examples of Spanish sentences from different sources are shown below:
\begin{itemize}
    \item \textbf{Written text in Wikipedia}: \textit{El español o castellano es una lengua romance procedente del latín hablado, perteneciente a la familia de lenguas indoeuropeas.} (Spanish or Castilian is a Romance language derived from spoken Latin, belonging to the Indo-European language
    family.)
    \item \textbf{Written text in customer support}: \textit{Mire, quería ver si me podían ayudar.} (Look, I wanted to see if you guys could help me)
    \item \textbf{Noisy ASR transcript in customer support}: \textit{Mire, este, es que, que- quería ver si me podían ayudar.} (Look, well, so I, I wanted to see if you could help me)
\end{itemize}

Recent advances in transformer-based pre-trained models have been proven successful in many NLP tasks across different languages. For Spanish, available pre-trained resources include multilingual models such as multilingual BERT (mBERT) \cite{devlin-etal-2019-bert} and XLM-RoBERTa (XLM-R) \cite{conneau-etal-2020-unsupervised}, as well as monolingual models such as BETO \cite{CaneteCFP2020}. However, large pre-trained models are trained on various written text sources such as Wikipedia and CommonCrawl \cite{cc:WenzekLachauxConneauChaudharyEtAl:2019:CCNet}, which are very distant from what we are trying to address in noisy ASR transcripts in the customer support domain. While Spanish is not usually considered a low-resource language in many NLP tasks, it is much more challenging to acquire sufficient training data in Spanish for our domain-specific task, since most of the publicly-available Spanish datasets do not come from natural human conversations, and have little coverage in the customer support domain. 

In addressing the challenge of in-domain data sparsity we make the following contributions:
\begin{enumerate}
\item We propose a punctuation restoration system dedicated for Spanish based on pre-trained models, and examine the feasibility of various pre-trained models for this task.
\item We adopt a domain adaptation approach utilizing out-of-domain Spanish text data.
\item We implement a data modification strategy and match in-domain English transcripts with Spanish punctuation usage, and propose a cross-lingual transfer approach using English transcripts.
\item We demonstrate that our proposed transfer learning approaches (domain adaptation and cross-lingual transfer) can sufficiently improve the overall performance of Spanish punctuation restoration in our customer support domain, without any model-level modifications.
\end{enumerate}

\section{Background}
\label{section:2}

Punctuation restoration is the task of inserting appropriate punctuation marks in the appropriate position on the unpunctuated text input. A variety of approaches have been used for punctuation restoration, most of which are built and evaluated on one language: English. The use of classic machine learning models such as n-gram language model \cite{Gravano} and conditional random fields \cite{lu-ng-2010-better} are common in early studies. More recently, deep neural networks such as Long Short-Term Memory (LSTM) \cite{HochSchm97} and transformers \cite{NIPS2017_3f5ee243} have been adopted in \cite{tilk15_interspeech} and \cite{courtland-etal-2020-efficient}. 

Punctuation conventions differ between Spanish and English. Namely, in addition to the equivalents of English and Spanish periods, commas, terminating question marks and terminating exclamation marks, we must also account for the inverted question marks (¿) and inverted exclamation marks (¡) used to introduce these respective clauses in Spanish. There has been limited work done in Spanish punctuation restoration and in most cases Spanish is covered as part of the multilingual training. \cite{li20m_interspeech} proposed a multilingual LSTM including the support for Spanish. \cite{PLN6377} uses a transformer-based model with both lexical and acoustic inputs for Spanish and Basque.

Transfer learning has been widely studied and applied in NLP applications for low-resource languages \cite{DBLP:journals/corr/abs-2007-04239}. Domain adaptation and cross-lingual learning both fall under the category of transductive transfer learning, where source and target share the same task but labeled data is only available in source \cite{ruder2019unsupervised}. Data selection is among the data-centric methods used in domain adaptation, which aims to select the best matching data for a new domain \cite{ramponi-plank-2020-neural}. \cite{fu-etal-2021-improving} uses data selection to improve English punctuation restoration with out-of-domain datasets. Recent advances in multilingual language models such as mBERT and XLM-R have shown great potential in cross-lingual zero-shot learning, wherein a multilingual model can be trained on the target task in a high-resource language, and afterwards applied to the unseen target languages by zero-shot learning \cite{hedderich-etal-2021-survey}. \cite{wu-dredze-2019-beto} and \cite{pires-etal-2019-multilingual} demonstrate the effectiveness of mBERT as a zero-shot cross-lingual transfer model in various NLP tasks, such as classification and natural language inference. 

\section{Methods}

\begin{figure*}[t]
\centering
\includegraphics[width=0.83\textwidth]{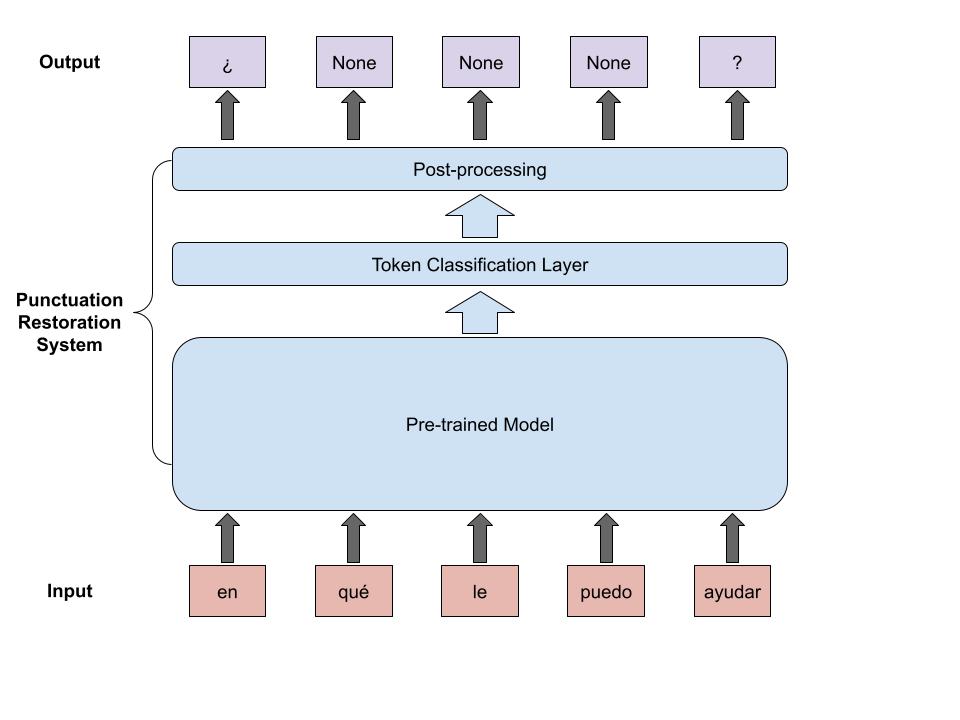} 
\caption{Our punctuation restoration system, showing the process of predicting “\textit{en qué le puedo ayudar}” as “\textit{¿En qué le puedo ayudar?}” (How can I help you?).}
\label{fig1}
\end{figure*}

\subsection{System Description}
\label{section:3.1}
Pre-trained transformer-based models have been widely adopted for various NLP tasks since the introduction of BERT \cite{devlin-etal-2019-bert}. Publicly available pre-trained models for Spanish include the multilingual models mBERT and XLM-R and the BERT-like monolingual model BETO. In this work, we evaluate all three pre-trained models in our experiments and compare their performance in both proposed domain adaptation and cross-lingual transfer approaches.

Using pre-trained models as a starting point, we formulate the Spanish punctuation restoration problem as a sequence labeling task, where the model predicts one punctuation class for each input word token. Instead of covering all possible Spanish punctuation marks, we only include nine target punctuation classes that are commonly used and important in terms of improving transcript readability:
\begin{itemize}
\item OPEN\_QUESTION: \textbf{¿} should be added at the start of this word token.
\item CLOSE\_QUESTION: \textbf{?} should be added at the end of this word token.
\item FULL\_QUESTION: \textbf{¿} and \textbf{?} should be added at the start and end of this word token respectively.
\item OPEN\_EXCLAMATION: \textbf{¡} should be added at the start of this word token.
\item CLOSE\_EXCLAMATION: \textbf{!} should be added at the end of this word token.
\item FULL\_EXCLAMATION: \textbf{¡} and \textbf{!} should be added at the start and end of this word token respectively.
\item COMMA: \textbf{,} should be added at the end of this word token.\footnote{The insertion of commas as decimal separators is not included here.}
\item PERIOD: \textbf{.} should be added at the end of this word token.
\item NONE: no punctuation should be associated with this word token.
\end{itemize}

The input to the Spanish punctuation restoration system is a transcribed utterance emitted by the ASR system. The ASR system outputs an utterance if an endpoint (long pause or speaker change) is detected in the audio. The length of a given utterance can vary, each utterance can contain multiple sentences, meaning that there can be multiple terminating punctuation marks -- period, question mark and exclamation mark -- in a single utterance.

The punctuation restoration model structure is illustrated in Figure \ref{fig1}. We add a token classification layer on top of the pre-trained models. Raw model prediction results are also post-processed by a set of simple heuristics to mitigate the error caused by unmatched predictions for paired punctuation marks. For instance, a predicted OPEN\_QUESTION class will be changed to NONE if there is no matched CLOSE\_QUESTION prediction in the same utterance. \footnote{This post-processing step may not always produce the correct result, but the overall prediction accuracy was improved by adding this post-processing in our experiments.}

\begin{table*}[t]
\centering
\begin{tabular}{| p{15cm} |}
\multicolumn{1}{|c|}{\textbf{Spanish out-of-domain (LDC) examples}} \\
\textit{Ah, qué bueno, yo conozco mucho cubano pero más que todo en Filadelfia.} (Ah, how good, I know many Cubans but especially in Philadelphia.)\\
\textit{Bueno, mira, eh, ¿sus papás, cuántos años llevan casados?} (Well, look, uhm, your parents, how long have they been married?)
\\
\hline
\multicolumn{1}{|c|}{\textbf{Spanish out-of-domain (OpenSubtitle) examples}} \\
\textit{Sé que lo que estoy pidiéndote es difícil.} (I know that what I’m asking you is hard.)\\
\textit{Sí, da un poco de tristeza.} (Yes, it makes you a little bit sad.)\\
\hline
\multicolumn{1}{|c|}{\textbf{Spanish in-domain examples}} \\
\textit{Buenas tardes, ¿cómo le puedo ayudar?} (Good afternoon, how can I help you?)\\
\textit{Pues no me funciona y lo he intentado varias veces.} (So, it doesn’t work and I’ve tried several times)
\\
\hline
\multicolumn{1}{|c|}{\textbf{English in-domain examples}} \\
\textit{I don’t find this app very helpful, I’m calling to cancel my subscription.}\\
\textit{Hi, this is Tom, how can I help you today?}
\\

\end{tabular}
\caption{Examples of Spanish and English utterances.}
\label{table1}
\end{table*}

\subsection{Datasets}
\label{section:3.2}

It is essential to acquire in-domain manual transcripts that come from real customer support scenarios to build a punctuation restoration model that fits the customer support domain. However, only around 5,000 in-domain transcribed Spanish utterances from call recordings could be obtained at this early product development stage. Additionally, there are around 200,000 in-domain manually transcribed English utterances from our call center product.

We supplemented this in-domain Spanish data with the Linguistic Data Consortium (LDC) Fisher Spanish Speech and Fisher Spanish Transcripts corpora \cite{Fisher-Spanish}. These corpora consist of audio files and transcripts for approximately 163 hours of telephone conversations from native Spanish speakers. These recordings are a good match to the acoustic properties of our telephone conversations, but the transcripts, which are mostly social calls with predefined topics, do not match the domain of customer support conversations. 

The Spanish portion of the OpenSubtitle corpus \cite{lison-tiedemann-2016-opensubtitles2016} also contains a variety of human-to-human conversation, albeit from movies rather than from spontaneous conversational speech. Spanish OpenSubtitle offers 179 million sentences from 192,000 subtitle files, and can provide our models with good exposure to exclamation marks, which are not included in the LDC dataset. However, the movie topics are generally distant from our business-specific, customer support domain. 

Some examples from both in-domain and out-of-domain data sources are illustrated in Table \ref{table1}. External out-of-domain datasets usually have various Spanish punctuation marks outside our supported range as described in \ref{section:3.1}. After reviewing the datasets from a linguistic perspective, we first apply a set of conversion rules to those unsupported punctuation marks without affecting the readability and semantic meanings: we delete quotation marks, replace colons and semicolons with commas, and replace ellipses with periods.

\subsection{Domain Adaptation}
\label{section:3.3}

Many machine learning applications have the assumption that training and testing datasets follow the same underlying distribution. But for our target task in the customer support domain, we mostly have to rely on external data such as LDC and Spanish OpenSubtitle during the training process, due to the lack of in-domain Spanish data. This will therefore cause a mismatch between our training and testing data in terms of its distribution, and consequently, performance will drop in our target task. Therefore, to mitigate this distribution mismatch, we apply domain adaptation on external Spanish datasets from two directions: data selection and data augmentation.

\begin{figure*} [t]

\centering
\begin{subfigure}[b]{0.45\textwidth}
\centering
\includegraphics[width=\textwidth]{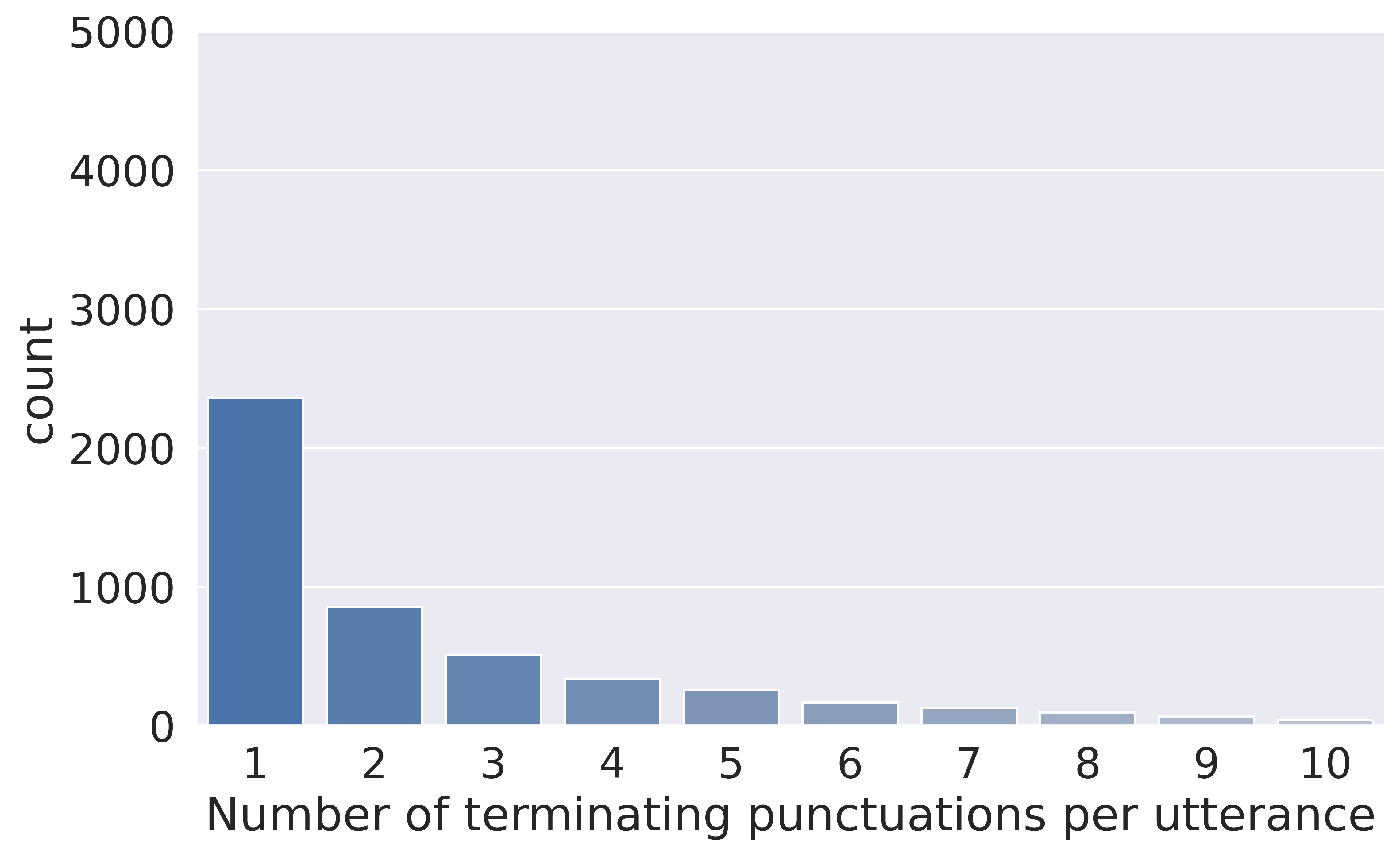}
\caption{in-domain}
\end{subfigure}
\medskip

\begin{subfigure}[b]{0.45\textwidth}
\centering
\includegraphics[width=\textwidth]{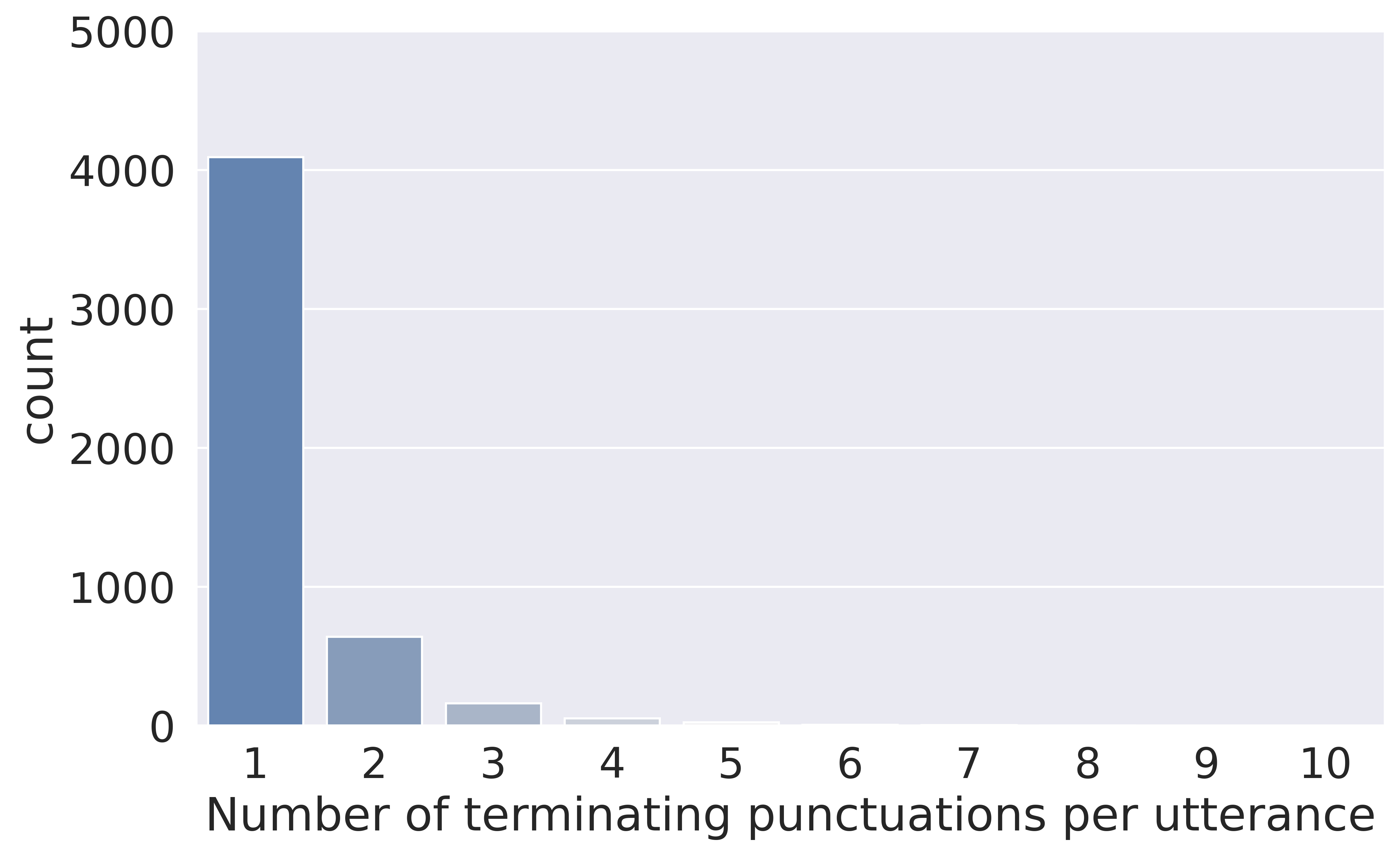}
\caption{LDC-before augmentation}
\end{subfigure}
\hfill
\begin{subfigure}[b]{0.45\textwidth}
\centering
\includegraphics[width=\textwidth]{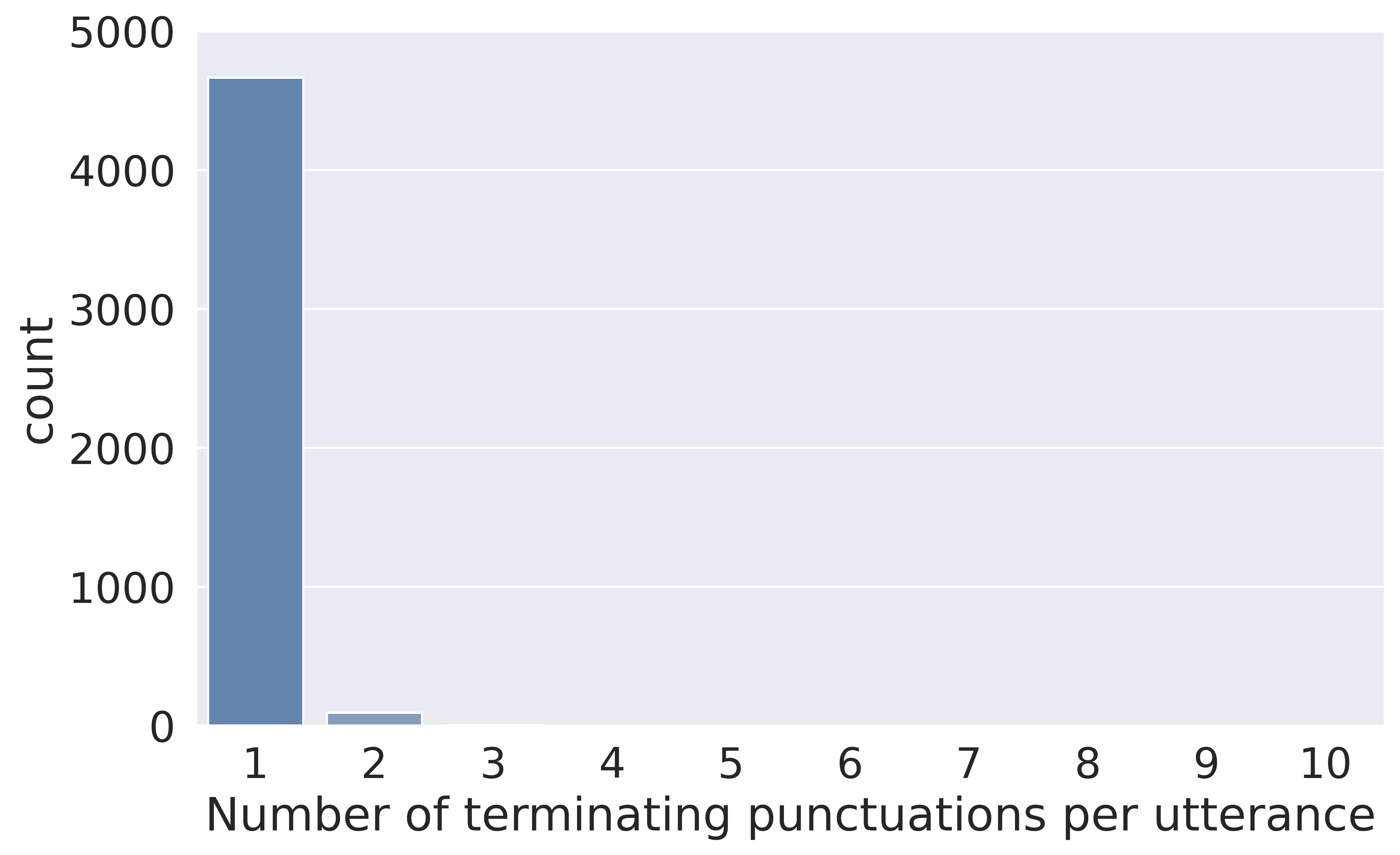}
\caption{OpenSubtitle-before augmentation}
\end{subfigure}
\medskip

\begin{subfigure}[b]{0.45\textwidth}
\centering
\includegraphics[width=\textwidth]{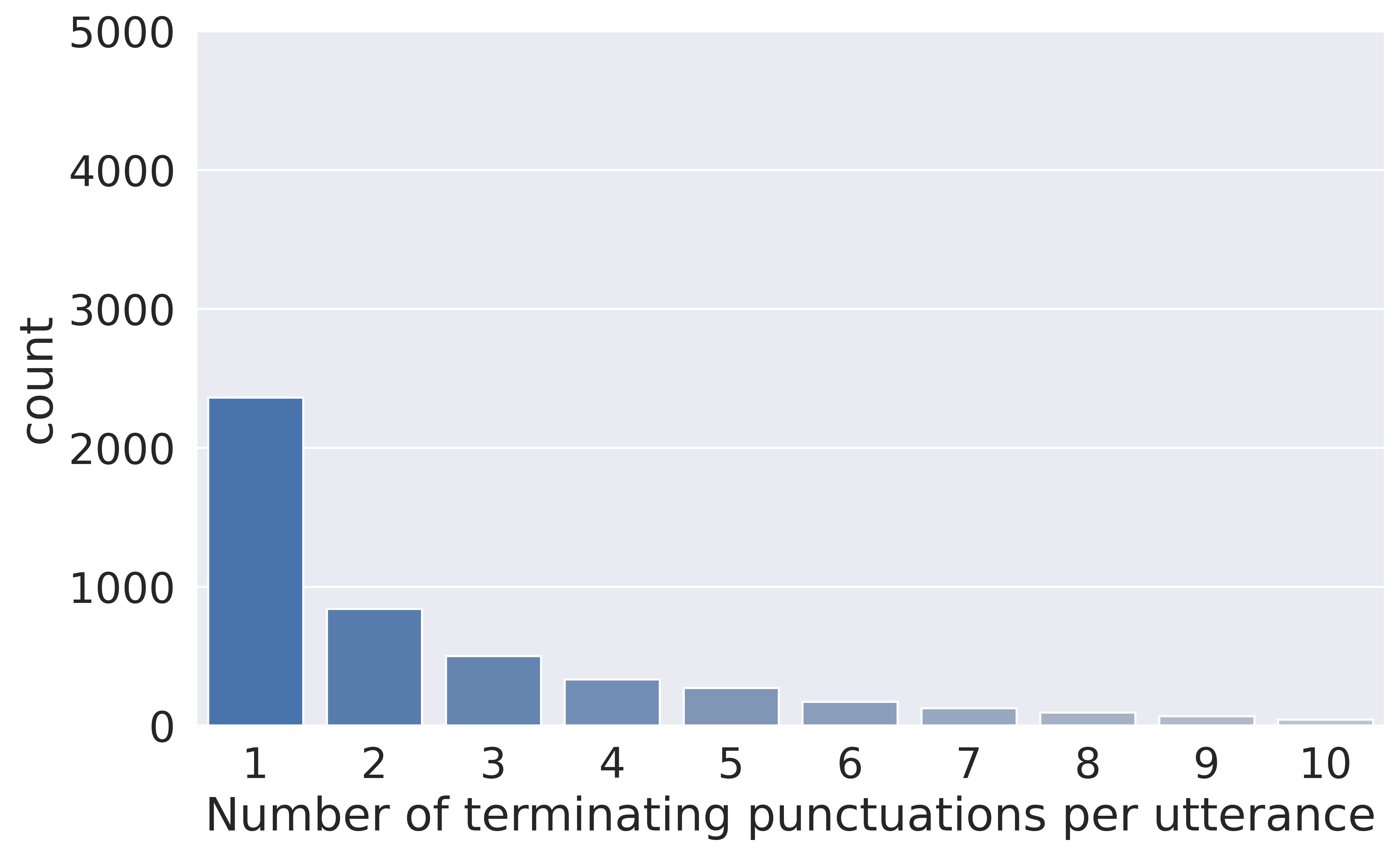}
\caption{LDC-after augmentation}
\end{subfigure}
\hfill
\begin{subfigure}[b]{0.45\textwidth}
\centering
\includegraphics[width=\textwidth]{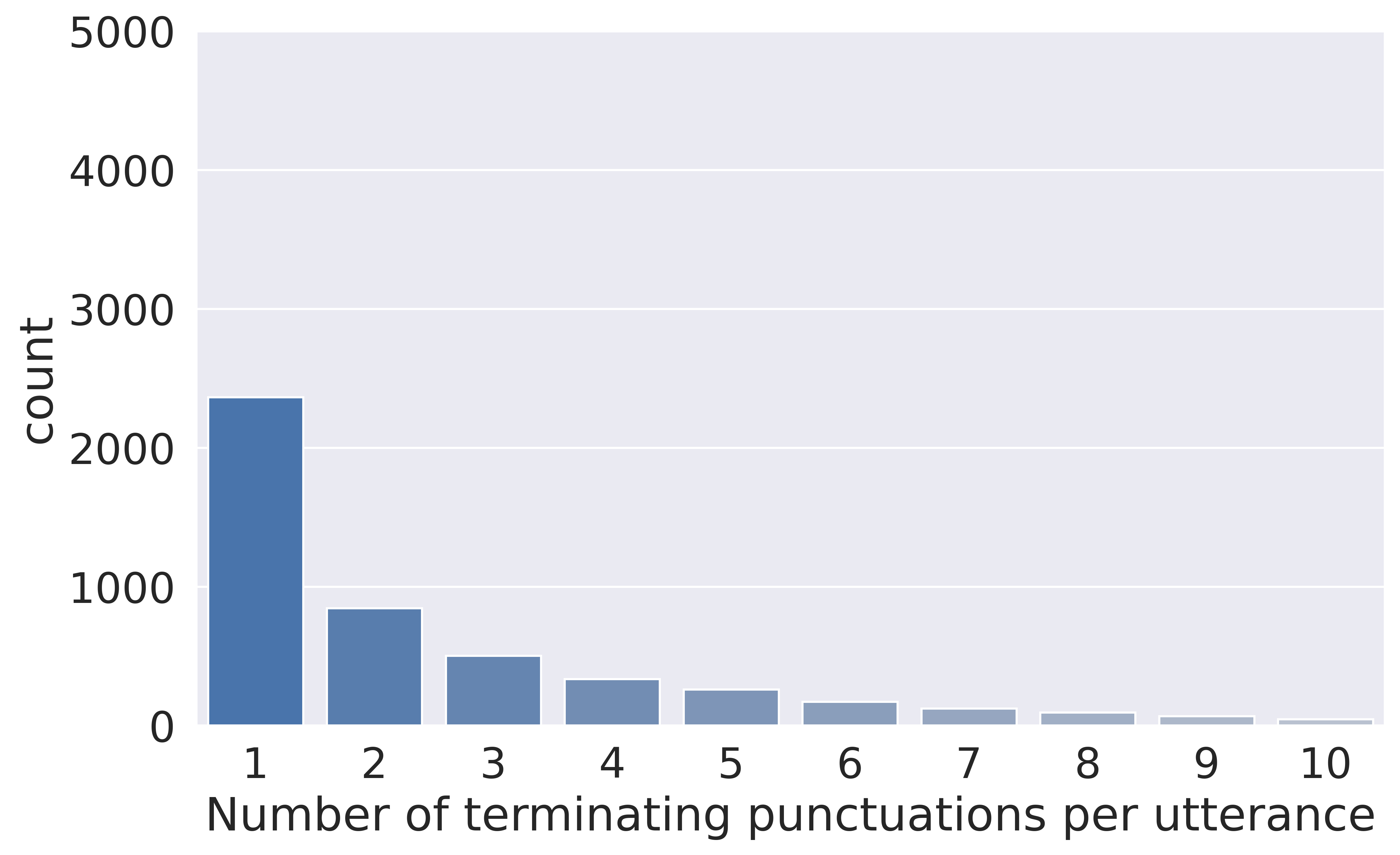}
\caption{OpenSubtitle-after augmentation}
\end{subfigure}

\caption{Comparison of number of terminating punctuations per utterance distribution in in-domain, LDC and OpenSubtitle datasets, before and after data augmentation.}
\label{fig2}
\end{figure*}

\subsubsection{Data Selection}
\label{section:3.3.1}

As described in \ref{section:3.2}, Spanish OpenSubtitle has a total of over 179 million sentences, which is much larger than our other data sources. However, the vast majority of the data in the Spanish OpenSubtile corpus are fundamentally distinct from our target customer support domain, and randomly sampling from out-of-domain datasets could hurt the model performance. Thus, following the procedure in \cite{fu-etal-2021-improving}, we first train a 4-gram language model using our Spanish in-domain data, and then sample the 100,000 utterances from the OpenSubtitle corpus with lowest perplexity (i.e. the highest language model similarity to the in-domain data).

Since the telephone conversation transcripts in the LDC corpora are closer to our target domain and there are only 130,000 utterances in this dataset, we do not perform further data selection on the LDC data for training purposes.

\begin{figure*}[t]
\centering

\begin{subfigure}[b]{0.45\textwidth}
\centering
\includegraphics[width=\textwidth]{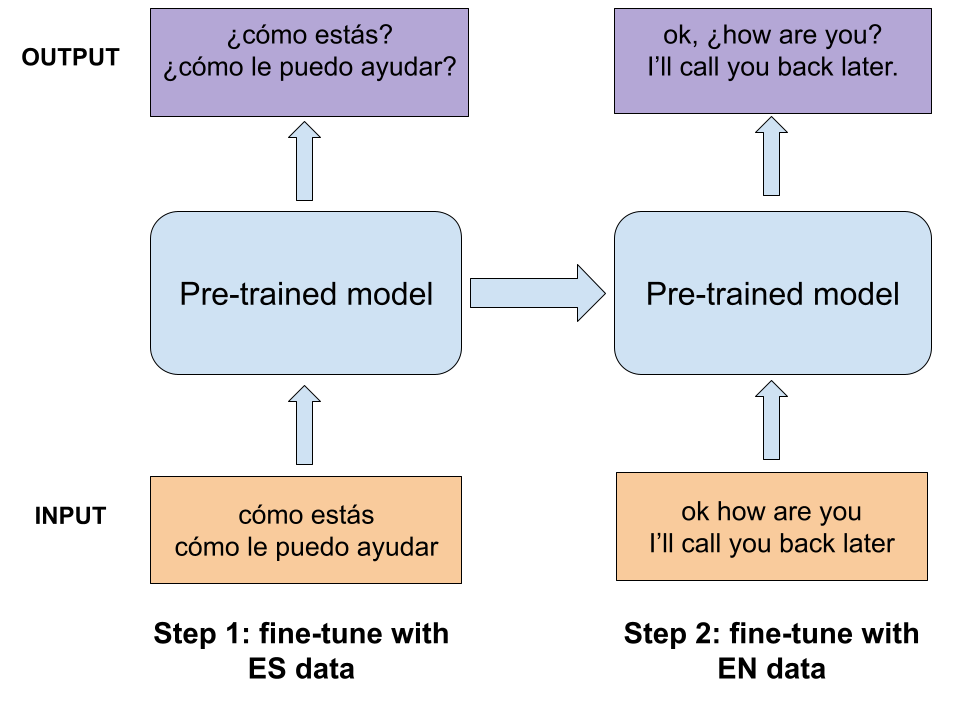}
\caption{}
\end{subfigure}
\hfill
\begin{subfigure}[b]{0.45\textwidth}
\centering
\includegraphics[width=\textwidth]{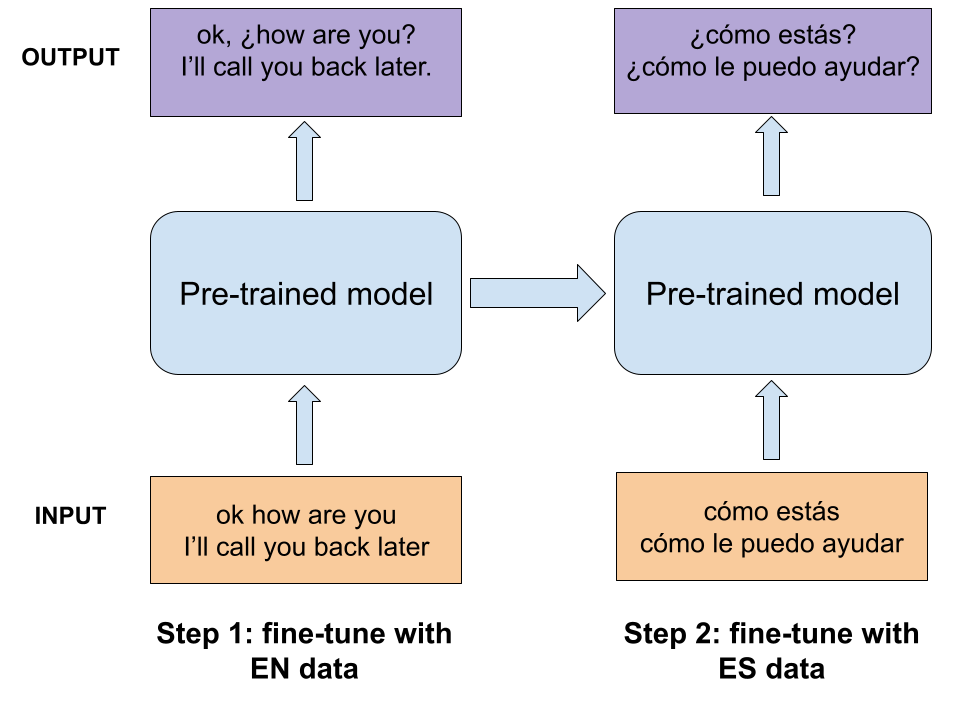}
\caption{}
\end{subfigure}

\medskip

\begin{subfigure}[b]{0.45\textwidth}
\centering
\includegraphics[width=\textwidth]{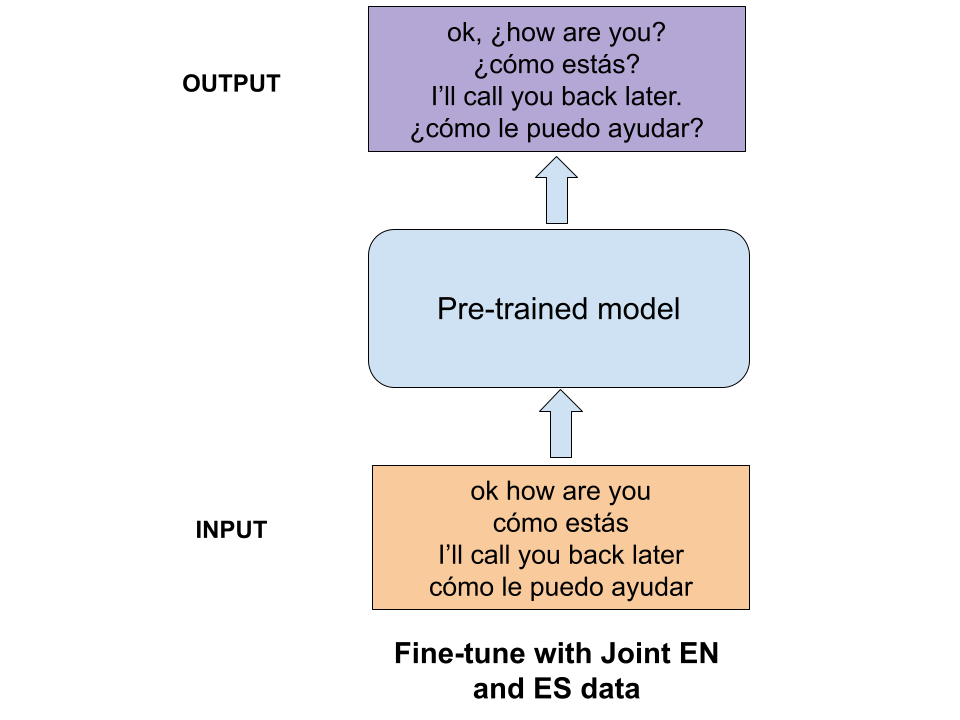}
\caption{}
\end{subfigure}

\caption{Diagram of three proposed fine-tuning strategies. (a) \textbf{ES->EN}, (b) \textbf{EN->ES}, (c) \textbf{Joint EN, ES}
}
\label{fig3}
\end{figure*}

\subsubsection{Data Augmentation}
\label{section:3.3.2}

Most of the data in LDC and OpenSubtitle datasets is segmented into single sentences. However, as described in \ref{section:3.1}, the input to our punctuation restoration system will be composed of larger blocks of utterances rather than single sentences. To illustrate this difference, we investigate how many terminating punctuation marks occur in each input from external datasets and in-domain data, respectively.

As shown in Figure \ref{fig2}(a)(b)(c), our in-domain data has a much wider distribution in terms of the number of terminating punctuation marks in a single utterance. However, the majority of samples in both LDC and OpenSubtitle consist of only one sentence each. It is necessary to augment the out-of-domain datasets to cover the wider spread of distribution exhibited in our in-domain data, based on the fact that this will affect how many terminating punctuation marks the model tends to predict per input utterance. We therefore apply data augmentation by concatenating sentences in these corpora, in proportion to the spread seen in our in-domain dataset, so that the overall terminating punctuation distribution in out-of-domain datasets matches our in-domain data. As Figure \ref{fig2}(d)(e) shows, the augmented results for the LDC and OpenSubtitle corpora more closely match the distribution of our in-domain Spanish data.

\subsection{Cross-lingual Transfer}
\label{section:3.4}

Multilingual language models such as mBERT and XLM-R advanced zero-shot cross-lingual transfer learning for low-resource languages \cite{hedderich-etal-2021-survey}. Instead of using cross-lingual transfer as zero-shot, we utilize our English in-domain data (described in \ref{section:3.2}) to fine-tune multilingual pre-trained models in addition to our available Spanish datasets to improve our Spanish punctuation restoration system. However, punctuation conventions differ between languages; to better leverage cross-lingual transfer learning, we first convert the punctuation usage in the source language to appropriately match the punctuation conventions in the target language. 

Since this study involves matching English punctuation to Spanish, the task is not insurmountable: most of the punctuation marks and their usages are the same across these two languages. Periods are used to terminate a declarative sentence in both languages, and the usage of commas to separate words or phrases is very similar. Therefore, no modifications are required for these two punctuation marks. 

One more significant challenge for this task is the fact that question marks and exclamation marks do work somewhat differently in Spanish writing than in English. Namely, in addition to the terminating role played in both languages by standard question marks (to denote the end of an interrogative sentence) and standard exclamation marks (to denote the end of an exclamatory sentence), Spanish writing conventions also require the addition of  an inverted question mark or an inverted exclamation mark, which occur at the beginning of the clause that contains the question or exclamation. For example:

\begin{itemize}
    \item \textbf{English}: \textit{Hi, how are you today?}
    \item \textbf{Spanish}: \textit{Hola, ¿cómo estás hoy?}
\end{itemize}

\begin{table*}[t]
\centering
\begin{tabular}{| c | c | c | c |}
\textbf{Training Data} &
\textbf{BETO} &
\textbf{mBERT} &
\textbf{XLM-R} \\

\textit{LDC} &
51.3\% &
50.2\% &
51.8\% \\

\textit{LDC + Selected OpenSubtitle} &
52.1\% &
51.5\% &
53.2\% \\

\textit{Augmented (LDC + Selected OpenSubtitle)} &
\textbf{53.7\%} &
\textbf{52.1\%} &
\textbf{54.7\%} \\
\end{tabular}

\caption{F1 score performance comparison using the LDC and OpenSubtitle datasets, before and after our domain adaptation approaches.}
\label{table2}
\end{table*}

For each question mark and exclamation mark in our English training data, we add an open question mark or exclamation mark, respectively, at the start of the word chunk that the terminating question or exclamation mark is in. 

For example, consider the following English utterance:

\textit{“OK, how can I help you?”}

For cross-lingual transfer training, it will be modified to:

\textit{“OK, ¿how can I help you?”}

By doing this conversion, the model will learn to predict punctuation as it should occur in Spanish contexts during the fine-tuning phase, even though what it actually sees are English utterances with Spanish punctuation. 

To determine the best way to transfer the in-domain distribution from English (\textbf{EN}) to Spanish (\textbf{ES}) in the punctuation restoration task, we investigate three fine-tuning strategies for cross-lingual transfer learning:

\begin{enumerate}
\item Fine-tune the pre-trained models in two steps, Spanish first then English. Noted as “\mbox{\textbf{ES->EN}}”. 
\item Fine-tune the pre-trained models in two steps, English first then Spanish. Noted as “\mbox{\textbf{EN->ES}}”.
\item Fine-tune the pre-trained models in one step, with joint English and Spanish data. Noted as “\mbox{\textbf{Joint EN, ES}}”
\end{enumerate}

Diagrams of three fine-tuning strategies are illustrated in Figure \ref{fig3}. Note that our objective is to build a model for Spanish, but it is still worth experimenting with “\mbox{\textbf{ES->EN}}” setting to establish the impact of more in-domain data albeit in a different language.

\section{Evaluation}
\subsection{Evaluation Setup}
\label{section:4.1}

We evaluate our proposed transfer learning approaches using the datasets described in \ref{section:3.2}. Using the model architecture shown in Figure \ref{fig1}, we fine-tune pre-trained models using various data combinations and fine-tuning strategies to demonstrate the effectiveness of our proposed approaches. Pre-trained models including both monolingual (BETO) and multilingual (MBERT and XLM-R) are explored and evaluated.

The Spanish punctuation restoration system is intended to operate in real-time so that customer-support agents can review prior information communicated by a customer and to provide the input to product features such as automatically retrieving information to assist the agent. As shown in \cite{fu-etal-2021-improving}, reducing the number of layers from deep pre-trained models does not significantly impact accuracy for the punctuation restoration task. To reduce the computation time during inference, we take only the first six layers from the pre-trained models as our starting point.

To evaluate the model accuracy in our target customer support domain, we split our in-domain Spanish manual transcripts into three parts: the training set (60\%), the validation set (10\%) and the test set (30\%). The Spanish in-domain training set is over-sampled to make the size comparable to the other datasets. The performance of every model is evaluated on the in-domain test set after being fine-tuned on various combinations of training sources and processes.

\begin{table*}[t]
\centering
\begin{tabular}{| c | c | c | c | c |}
\textbf{Training data and strategy} &
\textbf{BETO} &
\textbf{mBERT} &
\textbf{XLM-R} \\

\textit{ES only (no cross-lingual transfer)} &
62.8\% &
61.5\% &
62.9\% \\

\textit{ES->EN} &
N/A &
59.1\% &
60.7\% \\

\textit{EN->ES} &
N/A &
62.0\% &
63.5\% \\

\textit{Joint EN,ES} &
N/A &
62.4\% &
\textbf{64.4\%} \\

\end{tabular}

\caption{F1 score performance comparison with and without cross-lingual transfer. \textbf{\textit{ES}}: the combination of Spanish datasets including (1) Augmented (LDC + Selected OpenSubtitle) as described in Table \ref{table2}; (2) Spanish in-domain transcripts. \textbf{\textit{EN}}: English in-domain transcripts.
}
\label{table3}
\end{table*}

\begin{table*}[t]
\centering
\begin{tabular}{| c | c | c |}
\hline
\backslashbox{Gold}{Prediction}&
\textbf{CLOSE\_QUESTION} &
\textbf{PERIOD}\\
\hline
\textbf{CLOSE\_QUESTION} &
223&
106\\
\hline
\textbf{PERIOD} &
37&
2177\\
\hline

\end{tabular}
\caption{Confusion matrix of CLOSE\_QUESTION and PERIOD on test set, using best performing XLM-R in \ref{section:4.3}}
\label{table4}
\end{table*}

\subsection{Performance with Domain Adaptation}
\label{section:4.2}

We evaluate the F1 score performance before and after the domain adaptation approaches proposed in \ref{section:3.3}. Pre-trained models are fine-tuned using the combinations of LDC and selected OpenSubtitle datasets only, and then evaluated on our in-domain test set. The results are shown in Table \ref{table2}. Both data selection and data augmentation improve the overall F1 score performance for all three pre-trained models, which demonstrates the effectiveness of our domain adaptation approaches for the Spanish punctuation restoration task. Among three different models, XLM-R shows the best performance under this setup, and outperforms the monolingual BETO model after domain adaptation.

\subsection{Performance with Cross-lingual transfer}
\label{section:4.3}

To understand the effect of cross-lingual transfer, we use all the available data sources described in \ref{section:3.2}. We separate the Spanish datasets (LDC, selected OpenSubtitle and Spanish in-domain transcripts) from the English one (English in-domain transcripts), and fine-tune the pre-trained models using three different strategies described in \ref{section:3.4} (“\mbox{\textbf{ES->EN}}”, “\mbox{\textbf{EN->ES}}” and “\mbox{\textbf{Joint EN, ES}}”) as shown in Figure \ref{fig3}.

Table \ref{table3} shows our results on cross-lingual transfer learning: multilingual models (mBERT and XLM-R) both show performance gain with “\mbox{\textbf{Joint EN, ES}}” and “\mbox{\textbf{EN->ES}}” training. However, “\mbox{\textbf{ES->EN}}” training actually results in lower accuracy than models trained without cross-lingual transfer. As for the comparison with the monolingual model (BETO) which is not feasible for the direct cross-lingual transfer, XLM-R produces similar results as BETO without cross-lingual transfer, but XLM-R outperforms BETO by 1.5\% F1 score after joint training with both Spanish and English datasets. mBERT becomes comparable to BETO after cross-lingual transfer as well. 

\section{Future Work}
\label{section:5}
When analysing the prediction errors, we found that many CLOSE\_QUESTION classes are predicted as PERIOD by the model, as shown in Table \ref{table4}. This is a common behavior across all three pre-trained models, and is possibly due to the linguistic properties of Spanish. Because Spanish clauses do not require an overt subject noun phrase, and because Spanish has considerable variability in constituent order, it is often the case that there is no structural indication of whether an utterance should be interpreted as a declarative or as a question. Instead, intonation is used to make this distinction. For example, "\textit{hablan español}" ("they speak Spanish" or "do they speak Spanish") becomes a question with rising intonation. Future work in this area might focus on incorporating such acoustic information into punctuation restoration tasks.

\section{Conclusion}
\label{section:6}

For this study, we trained and tested a Spanish punctuation restoration system for the customer support domain based on pre-trained transformer models. To address in-domain data sparsity in Spanish, transfer learning approaches were applied in two directions: domain adaptation and cross-lingual transfer. We explored and fine-tuned three different pre-trained models with our transfer learning approaches for this task; our results demonstrate that the domain adaptation method improves the accuracy of all three pre-trained models. Cross-lingual transfer with joint training of English and Spanish datasets improves the performance of both multilingual pre-trained models. XLM-R substantially outperforms the monolingual BETO after cross-lingual transfer and achieves the best F1 score in our Spanish punctuation restoration task.

\bibliography{anthology,custom}

\end{document}